# Fast Sparse View Guided NeRF Update for Object Reconfigurations

Ziqi Lu*
Massachusetts Institute of Technology
ziqilu@mit.edu

Jianbo Ye*, Xiaohan Fei*, Xiaolong Li, Jiawei Mo, Ashwin Swaminathan, Stefano Soatto*
AWS AI Labs
{jianboye, xiaohfei, lxiaolx, Jiaweimo, swashwin, soattos}@amazon.com

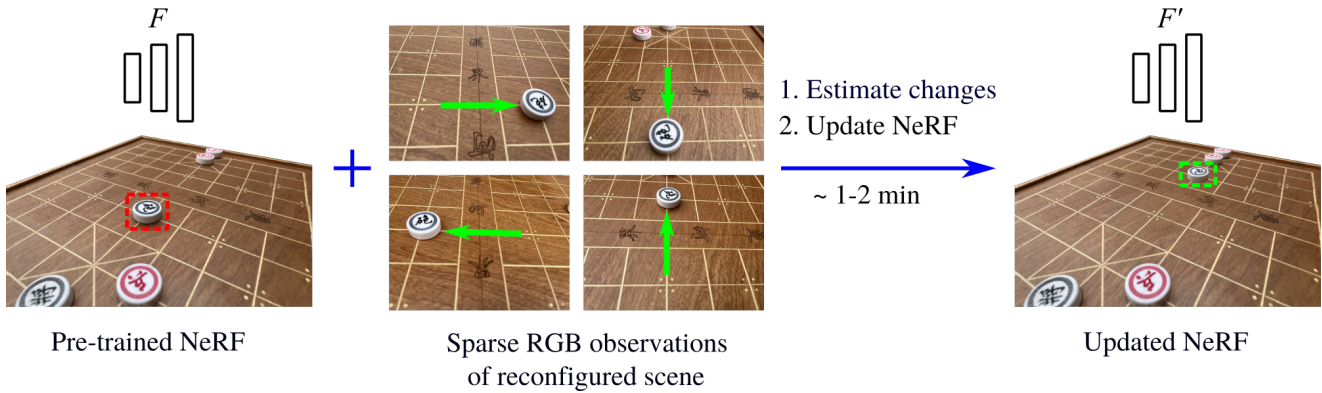

Figure 1. After training a NeRF for pieces on a Chinese chess board from hundreds of RGB images, a player moves the "Pao" piece, highlighted by the red box, to a new position. To accommodate this physical object reconfiguration, we take 4 additional images for the changed scene, and use our method to estimate scene changes and quickly update the pre-trained NeRF with the guidance of the 4 images.

## Abstract

*Neural Radiance Field (NeRF), as an implicit 3D scene representation, lacks inherent ability to accommodate changes made to the initial static scene. If objects are reconfigured, it is difficult to update the NeRF to reflect the new state of the scene without time-consuming data recapturing and NeRF re-training. To address this limitation, we develop the first update method for NeRFs to physical changes. Our method takes only sparse new images (e.g. 4) of the altered scene as extra inputs and update the pre-trained NeRF in around 1 to 2 minutes.*

*Particularly, we develop a pipeline to identify scene changes and update the NeRF accordingly. Our core idea is the use of a second helper NeRF to learn the local geometry and appearance changes, which sidesteps the optimization difficulties in direct NeRF fine-tuning. The interpolation power of the helper NeRF is the key to accurately reconstruct the un-occluded objects regions under sparse view supervision. Our method imposes no constraints on NeRF pre-training, and requires no extra user input or explicit semantic priors. It is an order of magnitude faster than re-training NeRF from scratch while maintaining on-par and even superior performance.*

## 1. Introduction

See Figure 1, two players are playing Chinese chess. Nerfy, a researcher working on Neural Radiance Fields (NeRFs) [32] approaches and asks: "May I build a 3D reconstruction for the pieces on your chessboard?" "Yes, please!" The players agree and pause their game. Nerfy takes out their smartphone and captures hundreds of images from various angles around the chessboard. The images are then uploaded to train a NeRF model with a state-of-the-art NeRF library (e.g. [20, 44]). The training takes about 30min to finish, after which Nerfy presents the NeRF's rendering results to the players. The players are amazed by its impressive novel view synthesis quality.

The players then resume their game. After moving a piece, one player asks Nerfy out of curiosity: "Can you update your NeRF to reflect the current layout of the game?" Nerfy responds "I can *redo* what I just did, capturing hun-dreds of images and training a new NeRF, but you will need to wait for another 30min". "Well, just move this piece in your NeRF to its current location." The player suggests.

---
*Work conducted during an internship at Amazon.

"No, ", Nerfy explains, "NeRF is an implicit 3D representation. It encodes the appearance and geometry of the scene with neural networks. It is difficult to know what parameters in the networks are responsible for this piece. Also, even if we can move this piece, the part of the board that is un-occluded by this move is also going to be missing from the new representation. So it is challenging to update the NeRF to reflect your move."

Motivated by the challenge Nerfy encounters, we developed a method to quickly update the pre-trained NeRF to reflect the object reconfigurations, and we only need to capture a sparse set of extra images for the changed scene. This method involves two primary steps: (1) estimating the scene changes and (2) updating the NeRF accordingly. The core idea of our method is to use a helper NeRF model to learn the geometry and appearance of the object un-occluded regions from scratch, sidestepping the optimization challenge of NeRF fine-tuning under the guidance of RGB images.

The major advantages of our method include: (1) No constraints on NeRF pre-training: Our method is agnostic of the NeRF model architecture and doesn't impose specific optimization requirements. We only need the pre-trained NeRF to have *reasonable* rendering quality at novel views. (2) No need for explicit semantic priors: We don't need category- or instance-specific training or semantic priors for the moved object. An object is considered as an object only if it is moved. (3) No extra user inputs: Our method doesn't require any additional user input, such as clicks or masks. It is designed to maximize users' convenience.

The major contributions of our work are: (1) A novel sparse-view-guided NeRF update training method. (2) A robust and efficient 3D change identification method for NeRF-represented scenes. (3) Experimental demonstration of our methods on self-collected real-world and simulated NeRF-update datasets.

Our method has broad real-world applications beyond chess playing. For instance, it is useful for large-scale NeRF 3D scanning and long-term NeRF-based robotic mapping. It can quickly accommodate local changes accidentally or intentionally made to the environment, eliminating the need for time-consuming image re-capturing and NeRF re-training. It also facilitates photo-realistic 3D scene monitoring from minimal camera setups and rendering dynamic physical objects in mixed reality.

## 2. Related Work

This section explores several other options for addressing the challenge Nerfy meets in Sec. 1.

### 2.1. Object-compositional NeRF

An obvious solution to the challenge of object re-configurations in NeRF is to train an object-compositional NeRF [11, 16, 17, 34, 45, 48, 50, 54]. These methods typically represent the scene as the composition of a background NeRF and multiple per-object NeRFs. After objects are re-configured, their respective NeRFs can be moved to their new locations to accommodate the scene changes. Nonetheless, this approach faces several limitations: (1) *Dependence on "pre-detected objects"*: Training such a NeRF requires pre-identification of objects through object segmentation networks. However, it is problematic to define an "object" before it is moved. For instance, if we pre-build an object NeRF for a "capped bottle", when the cap is removed from the bottle, simply moving the "capped bottle" NeRF cannot reflect the new state of the scene. (2) *Lack of supervision for un-occluded regions*: The areas unveiled by object movements receive no supervision during NeRF pre-training. As a result, their true geometry and appearance would be missing from the new representation after the object NeRFs are reconfigured. (3) *Constraints on NeRF architecture*: The scarcity of object-compositional NeRF implementations restricts their applications to more general NeRF uses cases, such as unbounded scene reconstruction. In contrast, our method identifies objects by detecting changes in the scene, reconstructs un-occluded areas under sparse view guidance, and does not impose limitations on the NeRF architecture.

### 2.2. Fast-training dynamic NeRFs

Another promising idea involves training dynamic NeRFs [7, 21, 22, 29, 36, 38, 42] from a RGB video capturing the scene change process. These methods typically train a 4D spatio-temporal NeRF or fit an extra deformation field to perform novel view synthesis at different time steps. Various methods [6, 12, 13, 26, 53] have also significantly accelerated the training of dynamic NeRFs, cutting down the training time from days to under an hour. However, many dynamic NeRF approaches make the assumption of temporal continuity, and may have a notable performance decline when scene dynamics are rapid or the video frame rate is low [8, 29]. In comparison, our method does not attempt to reconstruct the scene at every moment. It captures "3D snapshots" to represent the most up-to-date scene state and makes minimal assumption about how much the new sparse views overlap.

### 2.3. NeRF editting

NeRF-editing methods [25, 27, 33, 47, 49, 57, 58, 62] offer another solution to Nerfy's problem. These techniques typically utilize user inputs, such as clicks, drags, and masks to modify a pre-trained NeRF to reflect user instructions. A specific focus within NeRF editing [25, 33, 49, 57] is on object removal - reconstructing regions previously occupied by objects. This is often achieved by in-painting RGB and NeRF-rendered depth images, allowing for the fine-tuning or retraining of NeRF to eliminate unwanted objects. However, these methods are typically not optimized for training speed and the necessity for user edit instructions can undermine user convenience. On the contrary, our method is designed to quickly accommodate local appearance and geometry changes in a scene, guided by sparse observations of the changed scene. It requires no additional user input, offering greater flexibility across various scenarios.

## 3. Background

### 3.1. Neural Radiance Fields

NeRF is a 3D representation that implicitly encodes the geometry and appearance of a scene with a neural network: $F : (\mathbf{x}, \mathbf{d}) \rightarrow (\sigma, \mathbf{c})$. For each 3D point at position $\mathbf{x} = (x, y, z)$ observed from the viewing direction $\mathbf{d} = (\phi, \theta)$, NeRF predicts its RGB colors $\mathbf{c} = (r, g, b)$ and density $\sigma$. Given a calibrated camera with known pose $T_c^w \in \mathrm{SE}(3)$, we can perform volumetric rendering [14] to obtain the color at each image pixel. Along the camera ray $\mathbf{r}$ intersecting a pixel, we sample $n$ points and query NeRF to predict their colors $\mathbf{c}_i$ and densities $\sigma_i$. The predicted color at the pixel can be computed by volumetric integration as:

$$\hat{C}(\mathbf{r}) = \sum_{i=1}^{n} \exp(-\sum_{j=1}^{i-1} \sigma_j \delta_j)\left(1 - \exp\left(-\sigma_i \delta_i\right)\right) \mathbf{c}_i \quad (1)$$

where $\delta_i$ is the distance between consecutive point samples. At the NeRF training phase, the neural network $F$ is optimized by minimizing the RGB loss, measuring the differences between the predicted colors $\hat{C}$ and ground truth colors $C$ across training images $\{I_i\}_{i=1}^{N}$:

$$\mathcal{L} = \sum_{\mathbf{r} \in \mathcal{R}} \|\hat{C}(\mathbf{r}) - C(\mathbf{r})\|^2 \quad (2)$$

where $\mathcal{R}$ denotes the batch of rays sampled from the training images $\{I_i\}_{i=1}^{N}$.

### 3.2. Segment anything model

The segment anything model (SAM) [15] is a vision foundation model capable of predicting high-quality 2D segmentation masks for objects in an image from input prompts such as points and boxes. In this work, we use the box prompts to query SAM to obtain the predicted masks and prediction confidences on input images.

## 4. Methodology

Our objective is to rapidly update a NeRF to reflect physical scene changes, i.e. object re-configurations, with sparse-view guidance. As illustrated in Fig. 1, our method takes as input a NeRF $F$ pre-trained on dense RGB images $\{I_i\}_{i=1}^{N}$, and a sparse set of RGB images $\{I'_j\}_{j=1}^{M}$ observing the reconfigured scene. The output is an updated NeRF $F'$ representing the state of the scene after the object movements.

To fulfill this objective, we developed a pipeline to (1) identify scene changes and (2) update the NeRF accordingly. The scene changes, i.e. the middle results of our pipeline, as shown in Fig. 5, are represented by each rearranged object's (1) 3D segmentation mask $\mathcal{M} : \mathbb{R}^3 \rightarrow \{0, 1\}$ (for the object's previously occupied region); (2) 6D pose change $T \in \mathrm{SE}(3)$ [1] and (3) 2D segmentation masks $\{M_i\}_{i=1}^{N}$ on pre-training images $\{I_i\}_{i=1}^{N}$.

To maintain focus on the core idea, we will first elucidate the second stage of our pipeline: NeRF update (Sec. 4.1), prior to exploring the details of our scene change estimation method (Sec. 4.2). For brevity and ease of understanding, we describe our approach for the scenario of single-object pose change. However, our method can easily accommodate multi-object reconfigurations.

### 4.1. Fast NeRF update under sparse view guidance

As shown in Fig. 2, the object reconfiguration process can be considered as the composition of 2 separate processes: (1) relocate objects into their *move-in* regions; (2) remove objects from *move-out* regions. To accomplish the move-in, we transform the input coordinates for the pre-trained NeRF to account for the new object poses (Sec. 4.1.1). For move-out, we leverage a second helper NeRF to learn the un-occluded regions from scratch (Sec. 4.1.2).

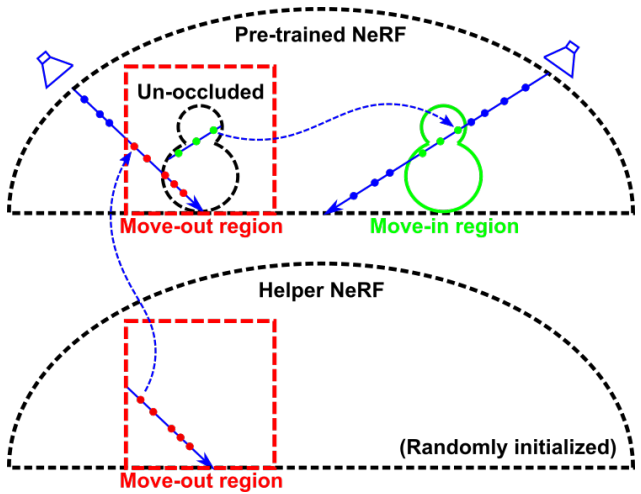

Figure 2. NeRF-Update: For the object *move-in* area, we estimate the object reconfigurations and *transform the NeRF input coordinates* to reflect the object movement. For the object *move-out* area, we leverage a *helper NeRF to learn the geometry and appearance of this region from scratch*.

---

[1]$T$ maps in-object points from their old positions to new positions. It can be interpreted as the 6D pose change of the object, when the object's canonical coordinate system is aligned with the world coordinate system.

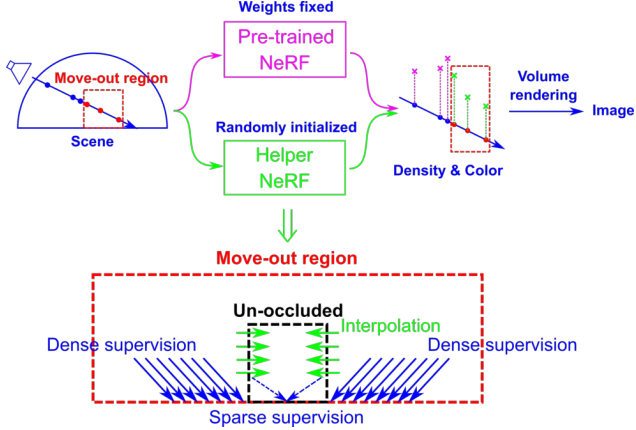

Figure 3. Core idea: Helper NeRF to learn local geometry and appearance changes for move-out regions in the scene. The *interpolation* power of the helper NeRF neural network is the key to accurately reconstruct the object un-occluded regions under *sparse* view supervision.

#### 4.1.1 Move objects into move-in regions

To move objects, we transform the coordinates $(\mathbf{x}, \mathbf{d})$ of the input point samples for the pre-trained NeRF $F$, to ensure that for any point samples falling within an object, we query NeRF at their prior locations and viewing directions preceding the object's pose change. This can relocate the object almost instantly without additional training.

In cases where an object with 3D mask $\mathcal{M}$ undergoes a pose change $T$, we can express the transformed input coordinates for NeRF as:

$$\tilde{\mathbf{x}}_T, \mathbf{d}_T^w = \begin{cases} T^{-1}\tilde{\mathbf{x}}, R^{-1}\mathbf{d}^w & \mathcal{M}(T^{-1}\tilde{\mathbf{x}}) = 1 \\ \tilde{\mathbf{x}}, \mathbf{d}^w & \text{otherwise} \end{cases} \quad (3)$$

where $\tilde{\ }$ denotes the homogeneous coordinates of 3D positions, and $R \in \mathrm{SO}(3)$ is the rotation matrix of the transformation $T$. It is important to note that the direct input viewing direction to NeRF, $\mathbf{d}^w \in \mathbb{S}^2$, is expressed in the world frame [32, 44]. We therefore directly transform the world-frame ray directions. For notation simplicity, nonetheless, we denote the pose-change-transformed input coordinates with $(\mathbf{x}_T, \mathbf{d}_T)$ in the following sections.

#### 4.1.2 Remove objects from move-out regions

Our object move-out method is designed based on two important intuitions: (1) **It is difficult to remove an object from a pre-trained NeRF's density field by direct fine-tuning**, especially with sparse RGB supervision (demonstrated in Sec. 4.1.3). In most cases, re-training a NeRF from scratch with the same data yields better reconstruction quality for the object un-occluded region. (2) **We are *not***

**solving a few-view NeRF reconstruction problem.** Our problem is not as challenging (demonstrated in Fig. 4(c) and Fig. 9(b)). Although each object un-occluded region only receives sparse supervision, the regions around it can receive abundant supervision from the pre-training images and thus can have good reconstruction quality. Provided there is a smooth transition between the two regions, we can leverage the *interpolation power* of neural networks to facilitate the NeRF reconstruction for the un-occluded region and to potentially guide the NeRF optimization to a good solution.

Therefore, in our method, we adopt a randomly initialized helper NeRF to learn the geometry and appearance of the object move-out regions from scratch [2]. Each object move-out region includes both the object un-occluded region and its adjacent areas. The helper NeRF is expected to leverage the dense supervision for the surrounding areas to aid the reconstruction of the object un-occluded region. The object move-out region in our implementation is represented with $\mathcal{M}^{\text{out}} : \mathbb{R}^3 \to \{0, 1\}$, which is computed by dilating the object 3D segmentation mask $\mathcal{M}$.

As depicted in Fig. 3, during volume rendering, we shoot rays $\mathbf{r}$ to the scene and sample $n$ points along each ray. For point samples falling out of the move-out area, we query the pre-trained NeRF to obtain density and radiance predictions. For samples within the move-out area, we instead query the helper NeRF. The volume rendering process in Eq. 1 can be re-expressed as:

$$\hat{C}(\mathbf{r}) = \sum_{i=1}^{n} \exp(-\sum_{j=1}^{i-1} \sigma'_j \delta_j)(1 - \exp(-\sigma'_i \delta_i)) \mathbf{c}'_i \quad (4)$$

where

$$\sigma'_j, \mathbf{c}'_j = \begin{cases} F^{\text{h}}(\mathbf{x}_j, \mathbf{d}_j) & \mathcal{M}^{\text{out}}(\mathbf{x}_j) = 1 \\ F(\mathbf{x}_{j,T}, \mathbf{d}_{j,T}) & \text{otherwise} \end{cases} \quad (5)$$

---

[2] To clarify, the helper NeRF is not "squeezed" into the move-out regions. It occupies the same 3D space as the pre-trained NeRF. Different regions in the helper NeRF can be allocated to reconstruct different object move-out regions.

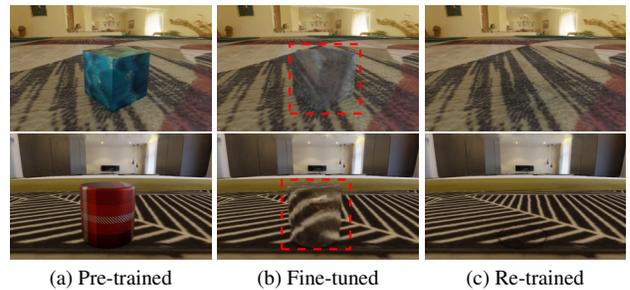

(a) Pre-trained    (b) Fine-tuned    (c) Re-trained

Figure 4. It is difficult to remove an object from a pre-trained NeRF by direct fine-tuning under sparse RGB supervision.

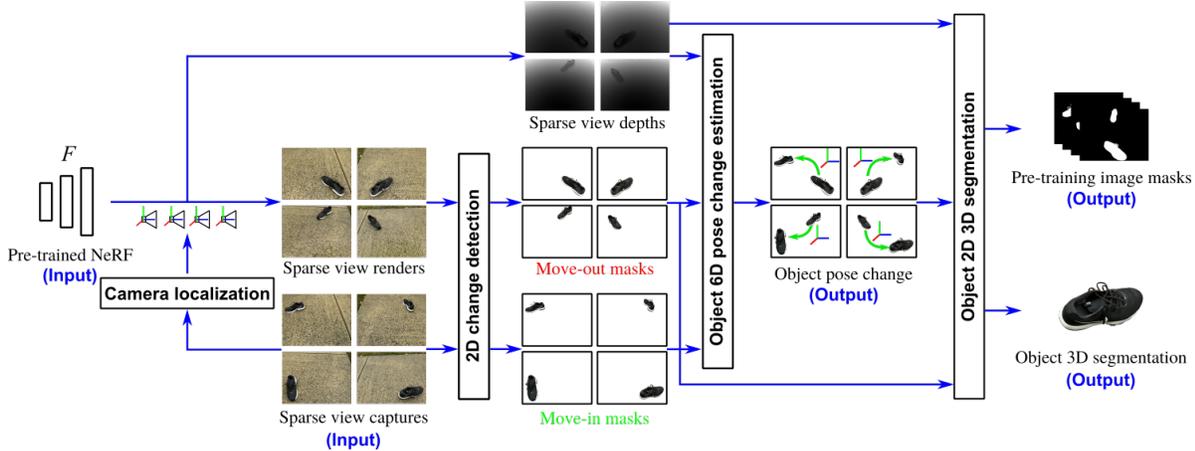

Figure 5. Scene change estimation pipeline

During the update training, we freeze the weights of the pre-trained NeRF $F$, and train the helper NeRF $F^h$ with the same RGB loss in Eq. 2. The training images, where the ray batch is sampled from, include both the sparse RGB images $\{I'_j\}_{j=1}^M$ and the pre-training images $\{I_i\}_{i=1}^N$. The object's previously and currently occupied areas are masked out from the pre-training images to avoid incorrect supervision. The unchanged parts that are far away from the move-out regions are also masked out from the pre-training images to focus the ray sampling around the un-occluded areas to speed up the training.

#### 4.1.3 No fine-tuning?! Why not?

We experimentally found that it is difficult to remove an object from a pre-trained NeRF's density field by direct fine-tuning. In most cases, we can obtain a better reconstruction performance by re-training a NeRF from scratch.

We conducted a series of experiments (same data setup as in Sec. 4.1.2) to study the feasibility of fast NeRF update through direct fine-tuning. Throughout these tests, we observed consistent challenges in altering the NeRF-encoded scene geometry (i.e., NeRF density field) by fine-tuning. For example, in the object removal tests illustrated in Fig. 4, when we attempt to fine-tune a object-centric NeRF to obtain a object-removed NeRF, there are always some remaining artifacts in the regions the object previously occupied, regardless of the adjustments in training hyper-parameters and epochs. Meanwhile, re-training the NeRF with the exact same hyper-parameters and data in most cases results in better reconstruction performance.

Similar findings were also reported in several studies on NeRF editing (e.g. [25, 57]). Specifically, NeRF-In [25] reported the difficulty of NeRF object removal by direct fine-tuning under the guidance of in-painted RGB images. They discovered that even with dense RGB image supervision, fine-tuning a pre-trained NeRF (baseline 1) cannot successfully remove unwanted objects from the NeRF density field. Re-training a new NeRF (baseline 2) usually yields more accurate scene geometry. These observations lend support to our findings at the dense-supervision regime.

Therefore, in our method, we sidestep the challenge of NeRF fine-tuning and use a helper NeRF to learn the geometry and appearance of unobserved regions from scratch.

### 4.2. Scene change estimation

This module's objective is to quickly and accurately estimate the changes in object configurations using the pre-trained NeRF $F$ and sparse-view observations of the re-organized scene $\{I'_j\}_{j=1}^M$. Built mostly on top of well-tested classic computer vision techniques and the foundation model SAM [15], our method makes minimal assumptions about the input and generalizes well across various objects and backgrounds. The modular design of our approach also allows for easy updates or replacements of specific sub-modules with alternative techniques. We summarize the procedure of our method as:

- Sparse view camera localization (Sec. 4.2.1)
- Sparse view change detection (Sec. 4.2.2)
- Object pose change estimation (Sec. 4.2.3)
- Object segmentation (Sec. 4.2.4)

#### 4.2.1 Sparse view camera localization

We first estimate the camera poses $\{T^w_{c'_j} \in \text{SE}(3)\}_{j=1}^M$ for the sparse-view images $\{I'_j\}_{j=1}^M$. We assume that before NeRF pre-training, the structure-from-motion pipeline (e.g. COLMAP [41]) would create a sparse point cloud of visual features, and store the camera calibration information. From these input, we use the HLoc library [40] to localize the cameras with the standard visual localization procedure.

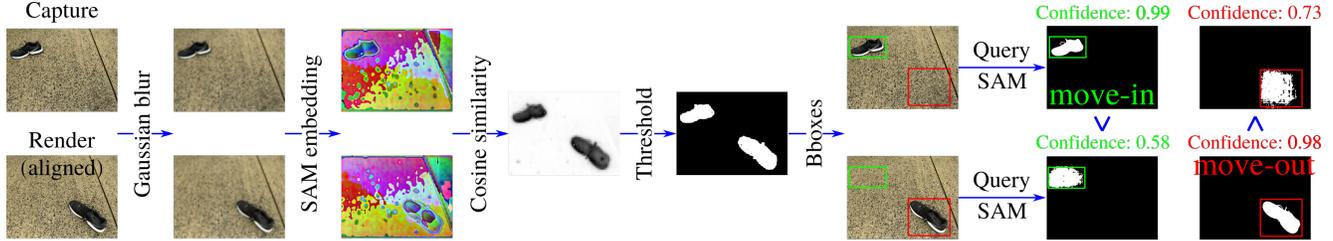

Figure 6. SAM-based 2D change detection for sparse views.

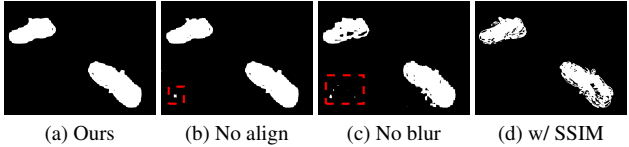

Figure 7. Threshold masks (before SAM query) with different variants of the proposed image differencing method. Our method is more likely to yield high-quality SAM predictions.

#### 4.2.2 Sparse view change detection

We propose a SAM-based image differencing method to identify 2D changes on the sparse-view images, as shown in Fig. 6. The input consists of the pre-trained NeRF $F$, sparse-view images $\{I'_j\}_{j=1}^M$ and their camera parameters. The output is the 2D move-in and move-out mask for each re-arranged object on each sparse-view image.

To begin with, we use the pre-trained NeRF to render RGB images at the sparse views: $\{\hat{I}'_j\}_{j=1}^M$. Note that the camera localization result from Sec. 4.2.1 is usually not pixel-perfect, especially in real world datasets. We therefore first align the renders and captures by homography estimation. We then Gaussian blur both images to filter out the high-frequency signals that are difficult to learn by the pre-trained NeRF. Afterwards, we extract the pixel-aligned SAM [15] feature maps from both set of images and compute the pixel-wise cosine similarity maps. We use Otsu's method [35] to threshold the maps and find contours that occupy substantial 2D areas. We then extract the 2D bounding boxes for the contours, and use them together with the captures and renders to query SAM. The move-in bounding box (green), for instance, incorporates the object on the captured image, but it only encloses some background pixels on the rendered image. We can therefore use the SAM prediction confidence to determine the type of the change mask: If the confidence for a SAM-predicted mask on the captured image is higher than that on the rendered, it is a move-in mask, otherwise, it is a move-out mask [3].

---
[3] We discuss the corner case of overlapped move-in and move-out masks in the supplementary material.

#### 4.2.3 Object pose change estimation

We developed an object pose change estimation and refinement method to compute the 6D transformation $T$ for each moved object. The input includes the pre-trained NeRF, captured images $\{I'_j\}_{j=1}^M$, object move-in and move-out masks and camera parameters for the sparse views.

We estimate the object pose changes by solving a variant of the perspective-n-point (PnP) problem. This initial estimate is refined through an analysis-by-synthesis approach, leveraging sparse-view RGB supervision to optimize $T$, similar to iNeRF [56]. Further details are provided in Sec. 9 of the supplementary material.

#### 4.2.4 Object segmentation

We further propagate the sparse-view object masks to all pre-training images to obtain 2D object masks $\{M_i\}_{i=1}^N$ on pre-training views and establish a 3D segmentation mask $\mathcal{M}: \mathbb{R}^3 \to \{0, 1\}$ for each pose-changed object. The input includes the pre-trained NeRF $F$, pre-training images $\{I_i\}_{i=1}^N$, the object move-out masks $\{M_j^{\text{out}}\}_{j=1}^M$ on sparse-view images, pose change $T$ and camera parameters.

To begin with, we use the pre-trained NeRF to render depth images at the sparse views: $\{\hat{D}'_j\}_{j=1}^M$, on which we apply the object move-out masks to obtain the object depth maps. We then use the camera poses $\{T_{c'_j}^w\}_{j=1}^M$ to convert the object depth maps to an object point cloud in the world frame. We project the object point cloud to all pre-training images and extract bounding boxes to query SAM to obtain move-out masks on all pre-training images.

We further propose a time- and memory-efficient "projection check" method to estimate the 3D object segmentation mask. First, we extract a 3D bounding box for each object point cloud. Second, within each 3D bounding box, we initialize a 3D binary occupancy grid. Third, we project every grid point to all the pre-training images, and check if the projections fall within the 2D object masks. If the projections for a grid point are within most 2D object move-out masks, the grid point is considered inside the object and evaluates to 1. At runtime, this object occupancy grid can be queried with arbitrary 3D positions in a very fast manner using trilinear interpolation. If the returned value is greater than $0.5$, the query point is considered inside the object. Since this "projection check" only involves (a manageable number of) matrix multiplication and mask reading operations, and the storage of sparse occupancy grids, it is much more time- and memory-efficient than its learning-based counterparts [9, 23, 28, 46, 59]. Also, as demonstrated in our experiments, our 3D segmentation results are sufficiently accurate to support NeRF-update.

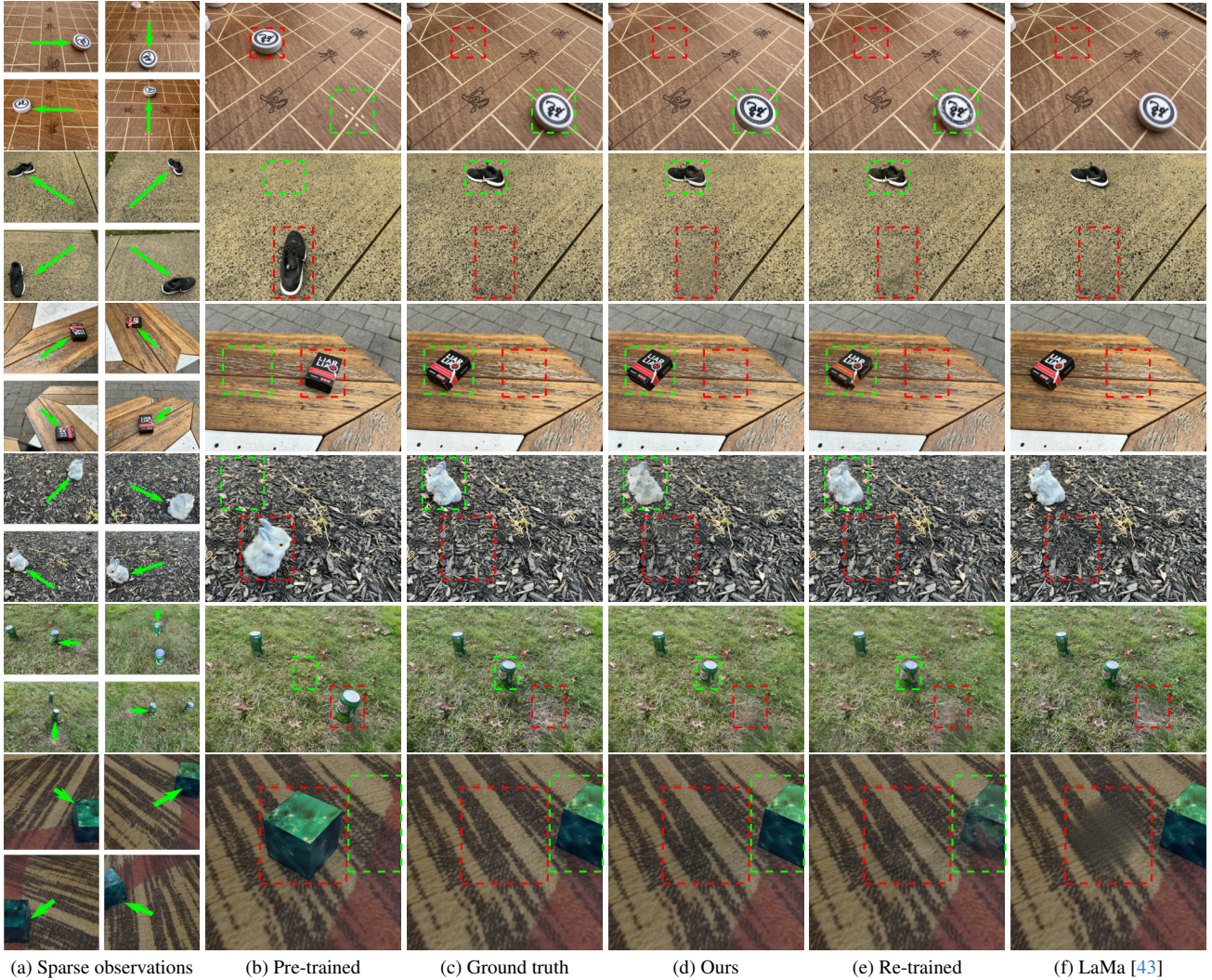

(a) Sparse observations  (b) Pre-trained  (c) Ground truth  (d) Ours  (e) Re-trained  (f) LaMa [43]

Figure 8. Qualitative results. (a) Sparse RGB observations guiding NeRF update; (b) Evaluation-view renders by pre-trained NeRFs; (c) Ground truth evaluation images; (d-f) Evaluation-view renders by our method and baselines. The move-in and move-out regions are highlighted with green and red boxes respectively.

Afterwards, we transform the object occupancy grid with the pose change $T$ and reproject its occupied grid points to obtain move-in masks on the pre-training images. We combine the move-in and move-out masks on the pre-training images to obtain the final segmentation masks $\{M_i\}_{i=1}^{N}$.

## 5. Experiments

We evaluate our method on a diverse set of real-world and synthetic data. Our method is developed on top of the NeRFacto model in NeRFStudio [44]. All experiments were conducted using a single NVIDIA Tesla V100 GPU. Please see the supplementary material for implementation details (Sec. 10) and full experimental results (Sec. 11).

### 5.1. Datasets

We collected 5 real-world datasets: Chess, Shoe, Liar, Bunny and Soda using an iPhone 13 mini. These data encompass a variety of objects, backgrounds and pose changes. In a typical data collection procedure, we first capture about 150 to 200 RGB images for an object-centric scene. We then move an object in the scene and capture 8 additional images for the changed scene from various angles. Out of the 8 new images, 4 are used for training, and the other 4 are used for evaluation.

Further, we generate 4 synthetic datasets: Cube, Cylinder, Sphere and Cone with Kubric [10], to evaluate the middle results of our pipeline against known ground truth values. [4]

### 5.2. Baselines

We propose to use the baseline of NeRF re-training. This baseline takes as input the pre-training images, sparse view new images, and the estimated object 2D segmentation masks from our method to mask out incorrect supervision on pre-training images, and re-train a new NeRF from scratch for the reconfigured scene.

We further propose to use the 2D in-painting model LaMa [43] as a second baseline to evaluate the reconstruction of the object *move-out* regions. This baseline takes as input the *evaluation images* and their object move-out masks, and outputs the masked pixels' colors. It is posited as a *reference* for the reconstruction *quality* of NeRF-in-painting methods for object-removal [25, 33, 49, 57], which without access to evaluation images, use similar in-painting models on training images to remove the target object and train an object-removed NeRF to render at evaluation views. Hence, we directly in-paint evaluation images for simple comparison with in-painting-based solutions.

### 5.3. Evaluation metrics

We employ PSNR, SSIM and LPIPS as the evaluation metrics. We compute these metrics separately for the 2D object move-in and move-out regions and denote them as PSNR/SSIM/LPIPS-in and PSNR/SSIM/LPIPS-out. To mitigate the camera localization errors for evaluation views in real data, we align the rendered images with the evaluation images prior to computation.

### 5.4. Results

The qualitative and quantitative analyses of our approach compared to the baselines are detailed in Fig. 8 and Tab. 1. Our method demonstrates high-quality reconstruction of the object *move-out* regions, which is *comparable* to the performance of the retraining baseline. This is *expected* since both methods reconstruct this region from scratch. For the *move-in* regions, however, our method considerably *outperforms* NeRF-retraining, since the latter only has 4-view observations of the object at its new location. Both our method and the retraining method outperform the LaMa baseline on PSNR-out and SSIM-out, because the in-painting method lacks direct observation of the reconfigured scene.

It's observed that our method does not outperform the retrain baseline as significantly in PSNR-in compared to in SSIM-in. This difference arises because PSNR, computed pixelwise, is more sensitive to misalignment between the ground truth and NeRF renders. Although we align the renders to captures prior to PSNR-in computation, the move-in regions on the two images may not be pixel-aligned. This can be confirmed with the results on the synthetic data, e.g., Cube, on which we have minimal misalignment and as a result our method's PSNR-in is also notably higher.

We further evaluate the runtime performance of our method. We show that our method finishes *within 2.5min* in most cases, which is an order of **magnitude faster** than the retraining baseline. If we only consider the training time, our update training is about **20 to 60 times faster** than NeRF re-training. Also, in scenarios involving multi-step NeRF-updates without new objects' movements, we can reuse the initial object segmentation, further *reducing the total runtime to less than a minute*. [5]

## 6. Ablation studies

We first study the effect of having object adjacent areas in the move-out region. As shown in Fig. 9(b), the reconstruction performance for the move-out region significantly degrades with a tight move-out region. We can discover remaining artifacts and unrealistic colors for the object un-occluded regions. This verifies our second intuition in Sec. 4.1.2. Fig. 9(c) illustrates the importance of the object pose change refinement module, without which the object poses are noticeably off. For full qualitative and quantitative results of the ablation studies, please see Sec. 12 in the supplementary material.

Further, we demonstrate our method's robustness to the *number* and *placements* of sparse-view cameras in Sec. 12 of the supplementary material.

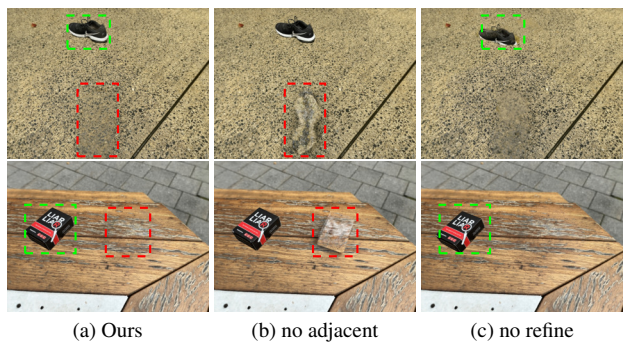

(a) Ours      (b) no adjacent      (c) no refine

Figure 9. Ablation study: Impact of (b) object-adjacent areas in move-out regions and (c) object pose change refinement on the updated NeRF's reconstruction quality.

---

[4]For more details about our data, please refer to Sec. 11.1 in the supplementary material for more details.

[5]Please refer to Sec. 11.3.2 in the supplement material for a runtime breakdown for different modules.

Table 1. Quantitative evaluation: PSNR-in (↑) and SSIM-in (↑) are computed for the move-in region. PSNR-out (↑) and SSIM-out (↑) are for the move-out region. "Total time (↓)" is the total NeRF update time including both the scene change estimation time and the "Train time (↓)". Please refer to Table 3 in the supplementary material for LPIPS values and full quantitative evaluation results.

|  | Methods | PSNR-in (↑) | SSIM-in (↑) | PSNR-out (↑) | SSIM-out (↑) | Total Time (↓) | Train time (↓) |
|---|---|---|---|---|---|---|---|
| Chess | Ours | **17.79** | **0.72** | **23.59** | **0.76** | **2.2**min | **49s** |
| Chess | Retrain | 15.83 | 0.65 | 22.91 | 0.67 | 21.5min | 20.2min |
| Chess | LaMa | - | - | 22.00 | 0.62 | - | - |
| Shoe | Ours | **17.27** | **0.56** | **16.97** | **0.44** | **2.5**min | **52s** |
| Shoe | Re-train | 16.25 | 0.39 | 16.94 | **0.44** | 23.5min | 21.9min |
| Shoe | LaMa | - | - | 13.47 | 0.36 | - | - |
| Liar | Ours | **18.49** | **0.80** | **22.03** | **0.76** | **2.5**min | **26s** |
| Liar | Retrain | 18.48 | 0.74 | 20.22 | 0.70 | 22.0min | 19.9min |
| Liar | LaMa | - | - | 17.81 | 0.59 | - | - |
| Bunny | Ours | **17.27** | **0.49** | 16.00 | 0.62 | **2.3**min | **46s** |
| Bunny | Retrain | 17.21 | 0.43 | **16.79** | **0.65** | 24.9min | 23.3min |
| Bunny | LaMa | - | - | 10.62 | 0.34 | - | - |
| Soda | Ours | **17.39** | **0.50** | 14.61 | **0.30** | **2.8**min | **63s** |
| Soda | Retrain | 15.92 | 0.39 | **14.83** | 0.27 | 21.8min | 20.1min |
| Soda | LaMa | - | - | 11.86 | 0.26 | - | - |
| Cube | Ours | **29.96** | **0.95** | 28.45 | **0.97** | **2.2**min | **22s** |
| Cube | Retrain | 18.46 | 0.68 | **28.76** | 0.96 | 21.8min | 20.0min |
| Cube | LaMa | - | - | 20.65 | 0.57 | - | - |

## 7. Limitations

Our approach does not reconstruct parts of reconfigured objects that are not visible in pre-training views. Also, it does not fully address complex environment lighting effects, including shadows and severe spatial lighting variations. Please refer to Sec. 13 in the supplementary material for a more in-depth discussion.

## 8. Conclusion

We develop a NeRF-update method that quickly reflects object reconfigurations under sparse view guidance. We show that our method is a working solution for a *real-world* challenge and it is a magnitude faster than NeRF retraining while achieving on-par and even superior reconstruction performance.

# 9. Object pose change estimation: Formulation

## 9.1. Background: Perspective-n-point problem

Perspective-n-point (PnP) is the problem of estimating the 6DoF pose of a calibrated camera from a set of 3D points and their corresponding 2D projections on the image. The perspective projection model for the set of 3D points $\mathbf{p} \in \mathbb{R}^3$ and their corresponding 2D projections $\mathbf{u} \in \mathbb{R}^2$ can be expressed as:

$$\pi\left(K T_w^c \tilde{\mathbf{p}}\right) = \mathbf{u} \tag{6}$$

where $\pi: \mathbb{R}^3 \to \mathbb{R}^2$ is the camera projection function, and $K$ and $T_w^c$ are the intrinsics and extrinsics for the calibrated camera. If we are given a set of 3D points $\mathbf{p} \in \mathbb{R}^3$, a set of 2D image points $\mathbf{u} \in \mathbb{R}^2$ and their correspondences $\mathcal{C}(\mathbf{u}, \mathbf{p})$, we can estimate the pose of the calibrated camera with a PnP solver $\phi$ (e.g., [18]) as:

$$T_w^c = \phi\left(K^{-1}\mathbf{u}, \mathbf{p}, \mathcal{C}(\mathbf{u}, \mathbf{p})\right) \tag{7}$$

## 9.2. Object pose change estimation

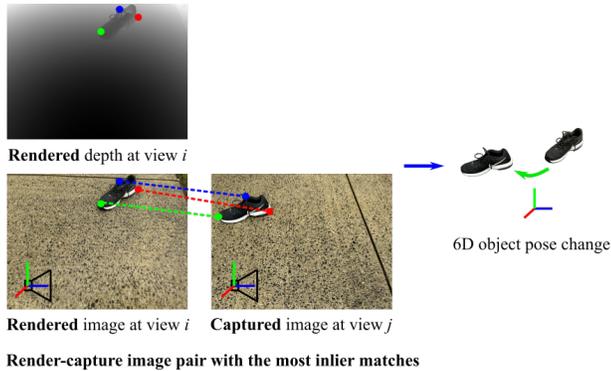

**Rendered** depth at view $i$

**Rendered** image at view $i$   **Captured** image at view $j$

**Render-capture image pair with the most inlier matches**

6D object pose change

Figure 10. Object pose change estimation

We formulate the object pose change estimation problem as a variant of the PnP problem. As shown in Fig. 10, we begin by detecting and matching the on-object visual features on the captured and rendered sparse-view images. Without knowledge about the object rotation, we don't know which image pair yields the most inlier matches, so we match features for all possible capture-render pairs, e.g. 16 pairs for 4 sparse views. We thus obtain the inter-image feature correspondences $\mathcal{C}(\mathbf{u}_i, \mathbf{u}'_j)$ for the object where $\mathbf{u}_i$ and $\mathbf{u}'_j$ represent the on-object features on the rendered image at view $i$ and the captured image at view $j$. For simplicity, instead of using all image pairs, we will use the image pair that has the most inlier correspondences to estimate the object pose change. Let's assume this image pair is the rendered image on view $i$ and the captured image on view $j$. For the captured image $j$, the detected keypoints $\mathbf{u}'_j$ are projections of on-object points after reconfiguration:

$$\pi(K_j T_w^{c_j} \tilde{\mathbf{p}}') = \mathbf{u}'_j \tag{8}$$

where $K_j$ and $T_w^{c_j}$ represent the camera intrinsics and extrinsics for sparse view $j$, and $\mathbf{p}'$ denotes the 3D positions of the on-object feature points after reconfiguration.

Note that we don't know $\mathbf{p}'$, the new 3D positions of on-object feature points. But we can transform their old positions with the pose change $T$, i.e., $\tilde{\mathbf{p}}' = T\tilde{\mathbf{p}}$, to obtain $\mathbf{p}'$ and replace it in Eq. 8:

$$\pi(K_j T_w^{c_j} T \tilde{\mathbf{p}}) = \mathbf{u}'_j \tag{9}$$

and the old positions of on-object points can be obtained by back-projecting $\mathbf{u}_i$, the 2D features on the rendered image $i$:

$$\tilde{\mathbf{p}} = T_{c_i}^w K_i^{-1} d_i \tilde{\mathbf{u}}_i \tag{10}$$

where $K_i$ and $T_w^{c_i}$ represent the camera intrinsics and extrinsics for sparse view $i$, and $d_i$ represent the NeRF-rendered depth values at the 2D feature positions. After substitution we can obtain:

$$\pi(K_j \underbrace{T_w^{c_j} T}_{} T_{c_i}^w K_i^{-1} d_i \tilde{\mathbf{u}}_i) = \mathbf{u}'_j \tag{11}$$

Comparing Eq. 11 and Eq. 6, we can first solve for the transformation $T_w^{c_j} T$ with a PnP solver and subsequently compute the pose change $T$ as:

$$T = T_{c_j}^w \phi\left(K_j^{-1} \tilde{\mathbf{u}}'_j, T_{c_i}^w K_i^{-1} d_i \tilde{\mathbf{u}}_i, \mathcal{C}(\mathbf{u}_i, \mathbf{u}'_j)\right) \tag{12}$$

## 9.3. Object pose change refinement

As demonstrated in Fig. 9(c), the initial object pose change estimate may be inaccurate, often due to feature matching errors, depth rendering noise, etc. We therefore refine the initial estimate with an analysis-by-synthesis approach.

Specifically, we freeze the weights of the pre-trained NeRF and use the pose-change-transformed input coordinates $\mathbf{x}_T, \mathbf{d}_T$ to perform volume rendering for on-object pixels at the sparse views:

$$\hat{C}(\mathbf{r}) = \sum_{i=1}^{n} \exp(-\sum_{j=1}^{i-1} \sigma_{j,T}, \delta_j)\left(1 - \exp\left(-\sigma_{i,T} \delta_i\right)\right) \mathbf{c}_{i,T} \tag{13}$$

where

$$\sigma_{i,T}, \mathbf{c}_{i,T} = F(\mathbf{x}_{i,T}, \mathbf{d}_{i,T}) \tag{14}$$

The color differences between the rendered and captured pixels are computed and the gradients of the residuals are back-propagated to optimize the initial $T$ estimate.

$$T = \underset{T \in \text{SE}(3)}{\arg\min} \sum_{\mathbf{r} \in \mathcal{R}} \|\hat{C}(\mathbf{r}) - C(\mathbf{r})\|^2 \tag{15}$$

For efficient optimization, we employ the same interest region sampling technique as iNeRF [56] to sample the ray batch $\mathcal{R}$.

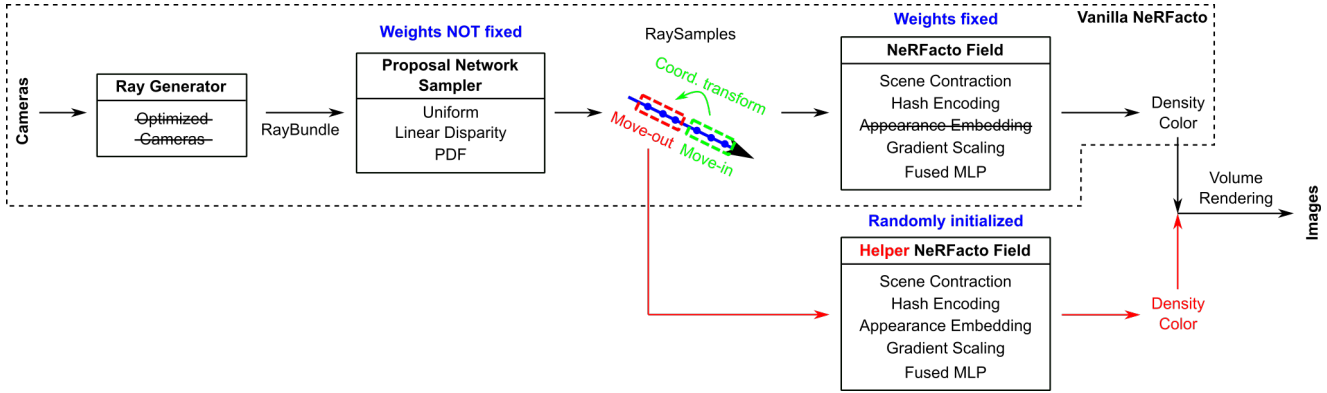

Figure 11. Implementation details of the updated NeRF model based on the NeRFacto model in NeRFStudio [44]

## 10. Implementation details

We present the implementation details of our NeRF update method in Sec. 10.1 and the scene change estimation pipeline in Sec. 10.2. Our method is mostly implemented with Pytorch 2.0 on top of the NeRFacto model in NeRFStudio [44] and tested on a NVIDIA Tesla V100 GPU.

### 10.1. Fast NeRF update under sparse view guidance

As illustrated in Fig. 11, our updated NeRF model builds upon the NeRFacto model [44]. From the input camera views, we first disable the camera pose optimizer in NeRFacto and generate ray bundles. Along these rays, we initially sample points with a piece-wise sampler [44]. These samples are then fed into a trainable proposal network sampler [2], containing two proposal density fields, to concentrate the samples in areas critical for volumetric rendering. Subsequently, we transform the 5D coordinates for the point samples (Sec. 4.1.1) to reflect the object pose change. These samples are input into the pre-trained NeRFacto field to get color and density predictions. Samples within the object move-out region, however, are input into a helper NeRFacto field (Sec. 4.1.2). The predicted colors and densities from the two fields are then combined to perform volumetric rendering to obtain the final image renders.

During update training, the weights of the pre-trained NeRFacto field are fixed, while the helper NeRFacto field is trained from scratch, and the proposal network is fine-tuned. Fine-tuning the proposal network is crucial for models with trainable proposal networks, as we discover this significantly aids in changing the geometry of the object move-out region, without notably affecting the reconstruction quality for the rest of the scene. Also, we disable the per-image appearance embedding [31] for the pre-trained NeRFacto field, avoiding the need to adjust the color MLP's input dimensions for new sparse-view images during NeRF update. Though, we enable the appearance embedding for the helper NeRFacto field to account for illuminance variations in training images. Furthermore, we employ the gradient scaling technique [37] to minimize floating artifacts in the pre-trained NeRF and the updated NeRF.

#### 10.1.1 NeRF pre-training

Before NeRF pre-training, we use the HLoc library [40] to perform structure from motion (SfM) on the pre-training images, to obtain the camera intrinsic and extrinsic parameters and a sparse point cloud of visual features. During the SfM process, we use SIFT [30] for feature detection and description, and Adalam [3] for feature matching. The sparse point cloud and the camera calibration results are saved to files for subsequent use in our scene change estimation pipeline (Sec. 10.2).

From the posed images, we pre-train a NeRF with the vanilla NeRFacto model. As previously mentioned, the camera pose optimizer and the appearance embedding are disabled in this stage. All other hyper-parameters are kept as their default settings. We train the NeRFacto model for 30000 iterations and it takes about 20∼30 mins to finish.

#### 10.1.2 Move objects into move-in regions

As described in Sec. 4.1.1, we map 5D coordinates for point samples in the move-in regions to their original coordinates to reflect object pose changes. However, for NeRF models using trainable proposal networks, we found it more effective to map the input coordinates both ways. This involves not only mapping points from move-in regions back to object previously occluded regions, but also mapping from the object un-occluded regions to the move-in regions. Such mapping re-initializes the proposal network for the move-out region, which would query the pre-trained proposal density field for the move-in region to propose input point samples. Since the move-in regions are typically vacant before object reconfigurations, their proposal density field closely resembles the current state of the move-out regions, fa-cilitating faster convergence of the proposal network fine-tuning. Without the two-way mapping, the proposal network would initially sample near the original object surface, leading to slower convergence. For an object with 3D mask $\mathcal{M}$ undergoing a pose change $T$, the transformed input coordinates for the pre-trained NeRFacto model can be re-expressed as:

$$\tilde{\mathbf{x}}_T, \mathbf{d}_T^w = \begin{cases} T^{-1}\tilde{\mathbf{x}}, R^{-1}\mathbf{d}^w & \mathcal{M}(T^{-1}\tilde{\mathbf{x}}) = 1 \\ T\tilde{\mathbf{x}}, R\mathbf{d}^w & \mathcal{M}(\mathbf{x}) = 1 \\ \tilde{\mathbf{x}}, \mathbf{d}^w & \text{otherwise} \end{cases} \quad (16)$$

One special case we need to be careful about though, is the potential overlap between the 3D move-in and move-out/object un-occluded regions. When such an overlap occurs, we take the overlapped region as the move-in region. The new move-out region is redefined as the original move-out region minus any overlapping move-in regions.

### 10.1.3 Remove objects from move-out regions

As mentioned in Sec. 4.1.2, we use a helper NeRF to learn from scratch the geometry and appearance of the object move-out regions. It is demonstrated in Fig. 9(c) that including object-adjacent areas in the move-out regions significantly improves the reconstruction of the object un-occluded regions. In our implementation, the object move-out region $\mathcal{M}^{\text{out}}$ is computed by dilating the object 3D segmentation mask $\mathcal{M}$. In principle, an infinitely large move-out region would provide enough data for interpolation and best aid the reconstruction of the un-occluded area. However, we observed that expanding the object 3D mask to be 2 to 3 times larger suffices for high-quality reconstruction of the un-occluded regions.

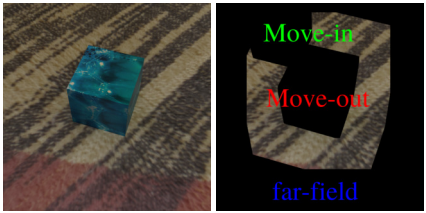

Figure 12. Example pre-training image and example masked pre-training image used for the NeRF update training.

The training data for NeRF update include both the pre-training images and the new sparse-view images. On the pre-training images, we apply three types of masks, object move-in mask, object move-out mask and far-field mask (for focused ray sampling), as shown in Fig. 12.

Masking the 2D move-in areas on pre-training images is critical for accurately reconstructing, counter-intuitively, the 3D *move-out* regions, This is because in some pre-training views, the 3D move-out region appear right in front of the 3D move-in region. Without masking the move-in areas, incorrect supervision would encourage the helper NeRF to form some cloud-like artifacts within the move-out regions to occlude the re-configured objects, making the 2D move-in areas resemble the background in these pre-training views.

The object move-out masks are predicted using SAM [15]. However, SAM may predict low-quality 2D object masks on some pre-training views, due to sub-optimal sparse view selections, camera pose errors, etc. To mitigate the influence of low-SAM-confidence ($< 0.95$) masks, we exclude the entire affected pre-training image from training. As long as these images are not the majority of the pre-training images (observed to be less than 3 per dataset in our experiments), we still have enough supervision signals from the rest of the pre-training views, to assist the reconstruction of object un-occluded regions.

Another challenge arises from the shadows near objects on the pre-training images, particularly in real-world data. If we directly use the predicted object move-out masks for training, due to NeRF interpolation, the shadows in the object-adjacent areas would make the un-occluded regions appear darker than their true appearance. To address this, we dilate the 2D move-out masks with a 3×3 kernel for $\lceil 0.01 \times W \rceil$ iterations, masking out the shadows. We found that this not only preserves but also enhances the reconstruction quality of un-occluded regions.

Additionally, to accelerate training, we apply a far-field mask on pre-training images, focusing ray sampling around the object move-out regions. This mask is obtained by dilating the object move-out mask with a 3×3 kernel for $\lceil 0.15 \times W \rceil$ iterations and unmasking only the newly dilated areas.

During the NeRF update training, the camera pose optimizer is turned off. The appearance embedding is disabled for the pre-trained NeRFacto field and enabled for the helper NeRFacto field. All other hyper-parameters remain at their default settings.

## 10.2. Scene change estimation

### 10.2.1 Sparse view camera localization

We employ the HLoc library [40] to localize the sparse-view cameras against the SIFT feature point cloud generated during SfM (Sec. 10.1.1). Camera calibration parameters are retrieved from the saved pre-training camera parameter files. The camera localization procedure involves (1) image retrieval with the NetVLAD model [1], (2) SIFT feature detection and matching with Adalam [3] (3) camera pose estimation and refinement with pycolmap [41].

### 10.2.2 Sparse view change detection

As illustrated in Fig. 6, our 2D change detection procedure involves (1) NeRF rendering, (2) image alignment, (3) Gaussian blur, (4) similarity computation, (5) thresholding, (6) bounding box extraction, (7) mask prediction and classification. The kernel size used for the Gaussian blur step is $\lceil 0.03 \times W \rceil$. At the bounding box extraction step, the contour area threshold beyond which we extract a bounding box is $0.01 \times H \times W$. After extracting the bounding box, we also slightly expand the box by 5% to improve the SAM prediction quality.

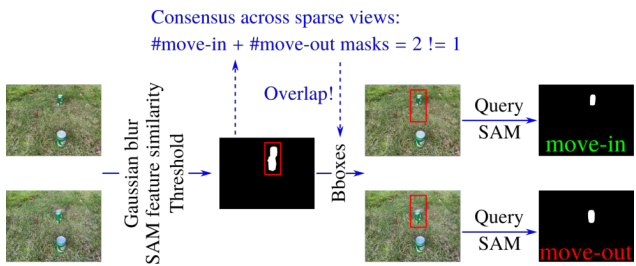

Figure 13. Corner case in change detection: overlapping 2D move-in and move-out areas on sparse views. We leverage the consensus across all sparse views to tackle the problem in our experiment on the Soda dataset.

A corner case we need to be careful about is the potential overlap of the 2D move-in and move-out masks on sparse views, as shown in Fig. 13. Typically, at the bounding box extraction step, we would obtain two separate bounding boxes for each object. However, such overlap could lead to one big bounding box incorporating both the move-in and move-out masks. Using this larger box to query SAM and identify mask types could cause us to miss one of the masks. While careful selection of sparse viewpoints can avoid such overlaps, we choose to directly address this challenge.

We tackle the problem leveraging the consensus across all sparse views. As long as such overlap doesn't happen on most sparse views, we can determine the accurate number of move-in plus move-out masks by taking the mode (most frequent count) of the numbers of threshold masks across all sparse views. In the cases of single object reconfiguration, for instance, the mode would be 2. If any view shows only 1 mask, we identify it as an overlap case. In such situations, we use the big bounding box to query SAM on both the captured and rendered images, without comparing their SAM confidences for mask classification, to obtain both the move-in and move-out masks.

In multi-object reconfiguration scenarios, we also face the challenge of object association, i.e., matching move-in and move-out masks belonging to the same object. We adopt the same object association method as used in the SAM paper (Sec. D.6 in [15]). Put it simply, we average-pool the SAM features within the move-in and move-out masks and use their cosine similarities to associate the masks.

### 10.2.3 Object pose change estimation

As discussed in Sec. 9, our object pose change estimation module comprises three major steps: (1) feature detection and matching, (2) pose change estimation and (3) pose change refinement.

For the first step, we use SuperPoint [5] for feature detection and description, and LightGlue [24] for feature matching. For pose change estimation, we utilize the absolute pose estimation method in pycolmap [41].

At the pose change refinement step, we reuse the forward pass functions for the NeRFacto model [44]. We uniformly sample 4096 rays within the interest regions [56] on the sparse-view images, which are created by dilating the detected on-object SuperPoint [5] pixels with a 3×3 kernel for 10 iterations. Along these rays, point samples are taken, their coordinates transformed, and then input into the pre-trained NeRFacto model for volumetric rendering and loss computation. Following iNeRF [56], we optimize the pose update variable $\delta T \in SO(3) \times \mathbb{R}^3$, which post-multiplies the initial $T$ estimate yields the current $T$ estimate. The pose update variable is initialized as zero rotation and translation. It is optimized with the Adam optimizer with a learning rate of 0.001 over 500 epochs, and an early stopping strategy is employed with a patience threshold of 100 iterations.

### 10.2.4 Object segmentation

As presented in Sec. 4.2.4, our object segmentation method involves three major steps: (1) move-out mask prediction on pre-training views, (2) 3D object segmentation, and (3) move-in mask computation on pre-training views.

For move-out mask prediction, we back-project object pixels on sparse views to form an object point cloud. Since the sparse-view camera poses or masks may not be pixel-perfect, there could be a small number of outliers in the computed point cloud. We remove the outliers by (1) computing the centroid position of the point cloud, (2) ranking the points by their distances to the centroid and (3) taking the nearest 95% as the object point cloud. We then project the filtered point cloud onto all pre-training views to derive 2D bounding boxes for SAM queries. To improve SAM prediction quality, these bounding boxes are expanded by 5% prior to SAM querying. We retain only SAM predictions with confidence above 0.95 for subsequent use. Pre-training images with low-confidence move-out mask predictions are excluded from both 3D object segmentation and the NeRF update training phases.

For 3D object segmentation, we begin by extracting a 3D bounding box around the object point cloud and expand-ing it to triple its original size. Within this expanded box, we initialize a 100×100×100 binary occupancy grid. Each grid point is projected onto the pre-training views to check if it falls within the object 2D move-out masks. If the projections fall within 95% of the move-out masks, the grid point would evaluate to 1, or otherwise it is set to 0.

Table 2. Data setup

| Datasets | Chess | Shoe | Bunny | Liar | Soda | Cube | Cylinder | Sphere | Cone |
|---|---|---|---|---|---|---|---|---|---|
| #Pretrain images | 136 | 161 | 151 | 202 | 174 | 180 | | | |
| #Sparse images | 4 training + 4 evaluation | | | | | | | | |
| Real/synthetic | Real | | | | | Synthetic | | | |
| Image resolution | 756×1008 | | | | | 1024×1024 | | | |

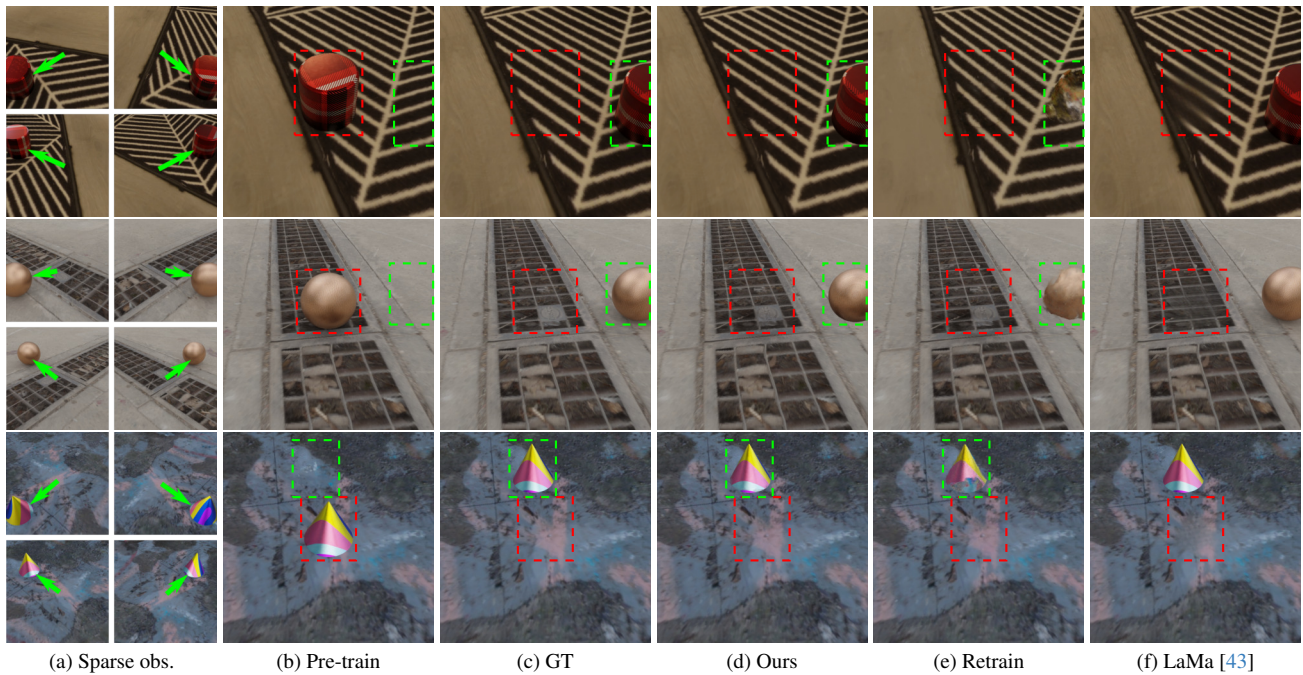

(a) Sparse obs.    (b) Pre-train    (c) GT    (d) Ours    (e) Retrain    (f) LaMa [43]

Figure 14. Additional qualitative results. (a) Sparse RGB observations guiding NeRF update; (b) Evaluation-view renders by pre-trained NeRFs; (c) Ground truth evaluation images; (d-f) Evaluation-view renders by our method and baselines. The move-in and move-out regions are highlighted with green and red boxes respectively.

For computing move-in masks, we project the occupied points from the occupancy grid onto the pre-training views. To fill interior gaps within the projected masks, we apply morphological closing using an 5×5 kernel for a single iteration.

## 11. Experimental details

### 11.1. Datasets

Additional details about our datasets are provided in Tab. 2, supplementing the information in Sec. 5.1.

For real data capture, we use an iPhone 13 mini mounted on a selfie stick, with its auto-focus and auto-exposure features disabled. The pre-training images were taken by pointing the camera from various angles towards the target objects, and trying to maintain a consistent camera-to-object distance. After the object is reconfigured, 8 new images are captured at about the same height from distinct angles, each approximately 45 degrees apart around the object's vertical direction. These raw images are subsequently downscaled 2 times to the resolution of 756×1006 for use in our method.

The synthetic datasets are generated with Kubric [10]. We place 180 pre-training cameras on a upper unit hemisphere, pointing inward to the target object at the center. These cameras are positioned at 10 different heights, and on each height the cameras are evenly distributed on a circular path. HDRI maps from `https://hdri-haven.com/` are used as the scene backgrounds, which are imported as textured domes into Kubric. The object texture maps are downloaded from `https://freepik.com`. The scene is lit with the ambient light (0.8, 0.8, 0.8), and we render RGB images at the resolution of 1024×1024. After we move the target object by the translation of (0.227, 0.227, 0), we render 8 additional images from cameras on the same height from viewing angles 45 degrees apart around the vertical axis of the object.

Table 3. Full quantitative evaluation of our method: PSNR-in (↑), SSIM-in (↑) and LPIPS-in (↓) are metric values for the move-in region. PSNR-out(↑), SSIM-out (↑) and LPIPS-out (↓) are for the move-out region. Note that the scene estimation and NeRF update training times are both considered in the computation of "Time (↓)".

| | Methods | PSNR-in (↑) | SSIM-in (↑) | LPIPS-in (↓) | PSNR-out (↑) | SSIM-out (↑) | LPIPS-out (↓) | Time (↓) |
|---|---|---|---|---|---|---|---|---|
| Chess | Ours | **17.79** | **0.72** | **0.16** | **23.59** | **0.76** | 0.23 | **2.2min** |
| | Retrain | 15.83 | 0.65 | 0.26 | 22.91 | 0.67 | 0.29 | 21.5min |
| | LaMa | - | - | - | 22.00 | 0.62 | **0.17** | - |
| Shoe | Ours | **17.27** | **0.56** | **0.21** | **16.97** | **0.44** | 0.24 | **2.5min** |
| | Re-train | 16.25 | 0.39 | 0.42 | 16.94 | 0.44 | 0.22 | 23.5min |
| | LaMa | - | - | - | 13.47 | 0.36 | **0.17** | - |
| Liar | Ours | **18.49** | **0.80** | **0.12** | **22.03** | **0.76** | 0.21 | **2.5min** |
| | Retrain | 18.48 | 0.74 | 0.20 | 20.22 | 0.70 | 0.26 | 22.0min |
| | LaMa | - | - | - | 17.81 | 0.59 | **0.18** | - |
| Bunny | Ours | **17.27** | **0.49** | **0.36** | 16.00 | 0.62 | 0.27 | **2.3min** |
| | Retrain | 17.21 | 0.43 | 0.51 | **16.79** | **0.65** | **0.25** | 24.9min |
| | LaMa | - | - | - | 10.62 | 0.34 | 0.31 | - |
| Soda | Ours | **17.39** | **0.50** | **0.33** | 14.61 | **0.30** | 0.53 | **2.8min** |
| | Re-train | 15.92 | 0.39 | 0.56 | **14.83** | 0.27 | 0.59 | 21.8min |
| | LaMa | - | - | - | 11.86 | 0.26 | **0.16** | - |
| Cube | Ours | **29.96** | **0.95** | **0.04** | 28.45 | **0.97** | **0.02** | **2.2min** |
| | Retrain | 18.46 | 0.68 | 0.39 | **28.76** | 0.96 | 0.03 | 21.8min |
| | LaMa | - | - | - | 20.65 | 0.57 | 0.29 | - |
| Cylinder | Ours | **24.40** | **0.83** | **0.09** | **32.24** | **0.96** | **0.02** | **2.5min** |
| | Retrain | 13.47 | 0.37 | 0.66 | 27.47 | 0.93 | 0.06 | 28.5min |
| | LaMa | - | - | - | 18.95 | 0.78 | 0.23 | - |
| Sphere | Ours | **23.82** | **0.70** | **0.23** | **27.47** | 0.94 | 0.06 | **2.2min** |
| | Retrain | 23.79 | 0.66 | 0.52 | 27.07 | **0.95** | **0.05** | 22.3min |
| | LaMa | - | - | - | 21.87 | 0.65 | 0.30 | - |
| Cone | Ours | **28.00** | **0.95** | **0.04** | **34.20** | **0.97** | **0.03** | **2.3min** |
| | Retrain | 13.14 | 0.72 | 0.38 | 34.00 | 0.96 | 0.04 | 22.0min |
| | LaMa | - | - | - | 24.46 | 0.85 | 0.21 | - |

### 11.2. Metrics

As mentioned in Sec. 5.3, we evaluate our method using PSNR, SSIM, and LPIPS, calculated separately for the 2D move-in and move-out areas on the evaluation images. The computation of PSNR-in and PSNR-out is straightforward, where we calculate PSNR pixel-wise within the respective masks. For SSIM and LPIPS, which are computed over entire images rather than pixelwise, we crop and resize the bounding box tightly around the move-in and move-out masks to a size of 256×256 to make sure each evaluation image contributes similarly to the final metric scores. We do not alter the unmasked pixels within these bounding boxes, such as setting them to all white or black, as doing so might bias the score computation.

### 11.3. Results

#### 11.3.1 Experimental results

Due to space constraints, only 6 out of 9 datasets' results are presented in Sec. 5.4. The remaining qualitative results are shown in Fig. 14. The full quantitative results are presented in Tab. 3. Our observations remain consistent: our method outperforms the retraining baseline in the move-in region reconstruction while achieving comparable performance in the move-out regions. An interesting observation is that although our method significantly outperforms the LaMa baseline on PSNR-out and SSIM-out, the LaMa baseline sometimes outperforms our method on LPIPS-out. This is likely because the deep-learning-based inpainting method is good at generating visually plausible and perceptually coherent fill-ins for missing parts of an image, potentially resulting in higher LPIPS, which measures perceptual image similarity.

In the additional results, we observe decreased performance of our method on the Sphere dataset. This is attributed to the change in the object's surface reflectance. Initially, the sphere reflects the black bricks on the ground, appearing darker. After reconfiguration, it reflects the grey ground, appearing brighter. Simply moving the sphere doesn't reflect the change in its appearance. This highlights a limitation in our method, as discussed in Sec. 13.2.

Table 4. Average runtime breakdown for modules in our pipeline.

| Modules | camera localiz. | change det. | pose change est. | object segment. | update training |
|---|---|---|---|---|---|
| Time | 4.1s | 13.6s | 10.7s | 87.5s | 43.5s |

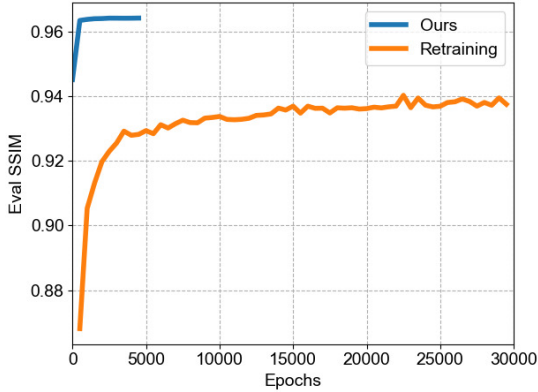

Figure 15. The evolution of evaluation SSIM for the update training stage of our method and the NeRF re-training baseline.

### 11.3.2 Runtime breakdown

In Table 4, we present the average runtimes for different components of our pipeline. For camera localization, HLoc takes about 4s to localize the 4 sparse-view cameras. For sparse-view change detection, the most time consuming steps are NeRF rendering (∼1 Hz) and SAM prediction (∼2 Hz). For pose change estimation, the initial estimation step is very fast, partially due to the use of the lightweight feature matcher LightGlue [24]. The refinement step is slower, taking about 9s to complete.

Object segmentation is the most time-consuming step, as it requires querying SAM (∼2Hz) for all pre-training images. However, there is potential for runtime improvements here, as object segmentation masks can be reused in multi-step object reconfiguration scenarios: After an object is moved the first time, we can cache its 3D mask, and skip object segmentation in future NeRF updates. If no *new* objects are moved in subsequent re-configuration steps, we can even skip both the object segmentation and update training steps, cutting the total runtime down to less than a minute. Additionally, using faster alternatives to SAM, e.g., [51, 60, 61], could further decrease the runtime.

Our NeRF update training typically converges within 500 to 1500 iterations, taking about 20s to 60s to finish. We determine the convergence for our method, and also for the re-training baseline, using SSIM computed across entire evaluation images. We prefer SSIM as it is less affected by image misalignment than PSNR and focuses more on the local patterns of the images than LPIPS. Training is considered convergent if the SSIM increase is less than

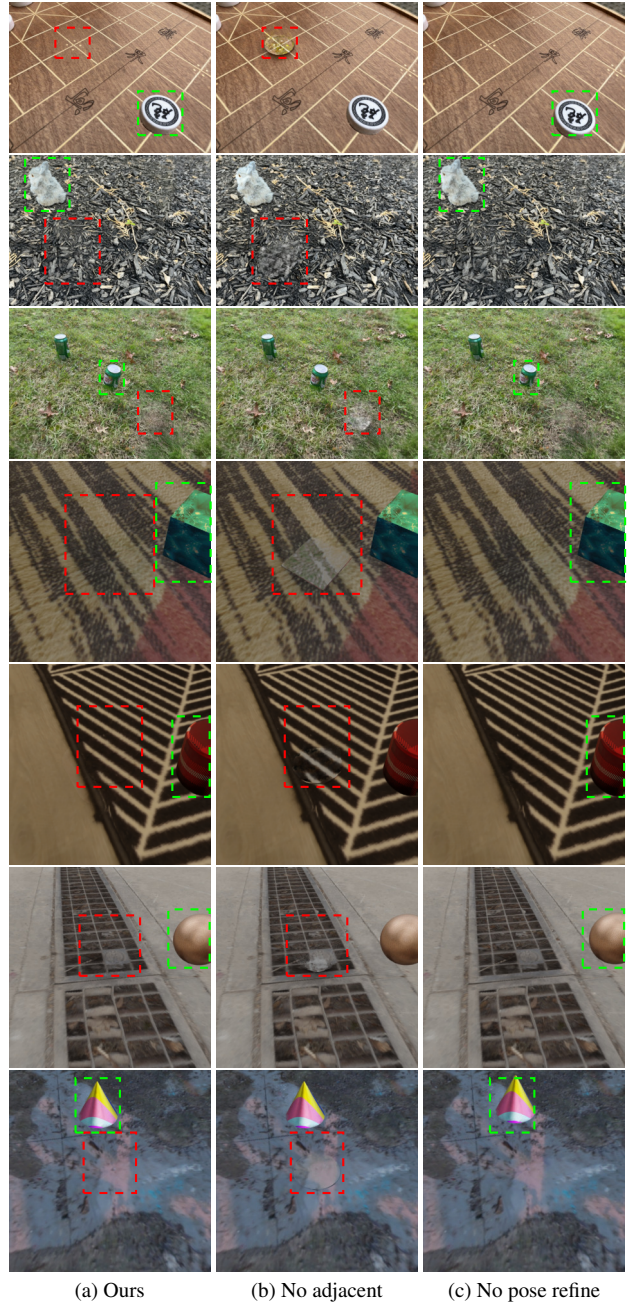

(a) Ours     (b) No adjacent     (c) No pose refine

Figure 16. Additional ablation study results: Impact of (b) object-adjacent areas in move-out regions and (c) object pose change refinement on the updated NeRF's reconstruction quality.

0.0005 over a 2500 iteration window. The training progress and SSIM evolution for the Cube dataset are illustrated in

Table 5. Full quantitative results of the ablation study. We study the impact of removing object pose change refinement module (×PR) and object-adjacent areas in move-out regions (×OA) on updated NeRF's reconstruction quality.

| Datasets | PSNR-in (↑) | | SSIM-in (↑) | | LPIPS-in (↓) | | PSNR-out (↑) | | SSIM-out (↑) | | LPIPS-out (↓) | |
| --- | --- | --- | --- | --- | --- | --- | --- | --- | --- | --- | --- | --- |
| | Ours | ×PR | Ours | ×PR | Ours | ×PR | Ours | ×OA | Ours | ×OA | Ours | ×OA |
| Chess | **17.79** | 8.25 | **0.72** | 0.33 | **0.16** | 0.39 | **23.59** | 18.79 | **0.76** | 0.53 | **0.23** | 0.45 |
| Shoe | **17.27** | 9.02 | **0.56** | 0.28 | **0.21** | 0.54 | **16.97** | 15.06 | **0.44** | 0.44 | **0.24** | 0.37 |
| Liar | **18.49** | 8.90 | **0.80** | 11.38 | **0.12** | 0.33 | **22.03** | 17.11 | **0.76** | 0.57 | **0.21** | 0.49 |
| Bunny | **17.27** | 13.21 | **0.49** | 0.32 | **0.36** | 0.49 | **16.00** | 14.35 | **0.62** | 0.52 | **0.27** | 0.34 |
| Soda | **17.39** | 9.80 | **0.50** | 0.23 | **0.33** | 0.61 | **14.61** | 14.57 | **0.30** | 0.30 | **0.53** | 0.55 |
| Cube | **29.96** | 24.12 | **0.95** | 0.81 | **0.04** | 0.08 | **28.45** | 25.14 | **0.97** | 0.89 | **0.02** | 0.17 |
| Cylinder | **24.40** | 21.46 | **0.83** | 0.69 | **0.09** | 0.11 | **32.24** | 27.33 | **0.96** | 0.89 | **0.02** | 0.14 |
| Sphere | **23.82** | 22.08 | **0.70** | 0.66 | **0.23** | 0.26 | **27.47** | 24.24 | **0.94** | 0.86 | **0.06** | 0.15 |
| Cone | **28.00** | 11.41 | **0.95** | 0.65 | **0.04** | 0.55 | **34.20** | 31.50 | **0.97** | 0.89 | **0.03** | 0.25 |

Fig. 15. For this dataset, our method converges within 500 iterations, which is 40 times faster than the retraining baseline, requiring about 20000 iterations.

### 11.3.3 Scene change estimation evaluation

We further evaluate the scene change estimation module of our method against the ground truth from the synthetic dataset. Specifically, the object pose change estimation errors are summarized in Tab. 6. We observe that most object pose change estimates are very accurate. All the estimation errors are much less than 1cm-1deg.

Table 6. Object pose change estimation errors on synthetic datasets. The size of the scene is about 2m×2m ×1m.

| Errors | Cube | Cylinder | Sphere | Cone |
| --- | --- | --- | --- | --- |
| Trans. | 0.05 cm | 0.06 cm | 0.11 cm | 0.05 cm |
| Rot. | 0.02 deg | 0.07 deg | 0.04 deg | 0.01 deg |

## 12. Ablation studies

### 12.1. Object-adjacent areas and object pose refinement

Additional ablation study results for removing object-adjacent areas and the pose change refinement step are presented in Fig. 16 and Tab. 5. Consistent with the findings in Sec. 6, incorporating the object-adjacent areas in the move-out regions is crucial for the reconstruction quality of the object un-occluded regions. Also, the object pose refinement step plays an important role in ensuring the accuracy of the objects' new poses.

### 12.2. Different number of sparse views

We assess the impact of changing the number of sparse RGB observations on our method's performance. We adjust the number of input sparse-view images in experiments with the Cube and Cone dataset to 1, 2, 3, 4, 8, 12, and 16, as shown in Fig. 17(a-e) and Fig. 18(a-e). We separately evaluate the effect on the scene change estimation step and NeRF update training step. For the scene change estimation step, we evaluate how a reduced number of sparse views influences the accuracy and potential failure of scene change estimation. For the update training step, we examine whether the changed amount of supervision from sparse views could significantly impact the reconstruction quality of the object un-occluded regions.

#### 12.2.1 Impact on scene change estimation

We use the same pre-trained NeRF and a varying number of sparse-view RGB images as input to the scene change estimation pipeline, and report its object pose change estimation errors in Tab. 7. With fewer sparse views (2- and 3-view), the pose change estimation accuracy slightly degrades, likely because of less supervision available for pose refinement. Nonetheless, the estimation remains highly accurate, with errors significantly below 1 deg-1 cm, thanks to the robustness of our pose change estimation module that makes minimal assumption about the input. However, our method fails in the single-view case, where our method fails to recover the full object point cloud from the NeRF-rendered single-view depth map, and as a result, object segmentation fails due to incorrect SAM box prompts on many pre-training images.

Table 7. Our scene change estimation pipeline is tested with different number of input sparse views. We report the object translation and rotation change estimation errors. The errors increase slightly with fewer views. With a single view, our method fails at the object segmentation step due to incomplete object point cloud.

| | Cube | | Cone | |
| --- | --- | --- | --- | --- |
| | Trans. error | Rot. error) | Trans. error | Rot. error |
| 1-view | Fail | Fail | Fail | Fail |
| 2-view | 0.039 cm | 0.139 deg | 0.016 cm | 0.154 deg |
| 3-view | 0.024 cm | 0.067 deg | 0.018 cm | 0.123 deg |
| 4-view | 0.020 cm | 0.053 deg | 0.012 cm | 0.055 deg |
| 8-view | 0.015 cm | 0.014 deg | 0.015 cm | 0.049 deg |
| 12-view | 0.014 cm | 0.015 deg | 0.011 cm | 0.056 deg |
| 16-view | 0.014 cm | 0.013 deg | 0.013 cm | 0.045 deg |

### 12.2.2 Impact on NeRF update training

As discussed in Sec. 4, the sparse-view supervision and interpolation from object-adjacent areas *jointly* ensure the accurate reconstruction of object un-occluded regions. We have demonstrated the critical role of interpolation in NeRF update training through ablation studies (See Fig. 16(b) and Tab. 5). Now we are curious whether the update training is sensitive to the amount of sparse-view supervision. Specifically, does changing the number of sparse views drastically influence the reconstruction quality?

To isolate the impact of the number of views from errors in the scene change estimation, we skip the estimation step and utilize the ground truth middle results as the input for NeRF update training. The qualitative and quantitative results of our method are reported in Fig. 17 and Fig. 18. From the results, we only observe small changes in PSNR-out (¡1dB), SSIM-out ($<<0.1$) and LPIPS-out ($<<0.1$) across different number of views. This indicates that our NeRF-update training method is robust to the variation of the number of input sparse views.

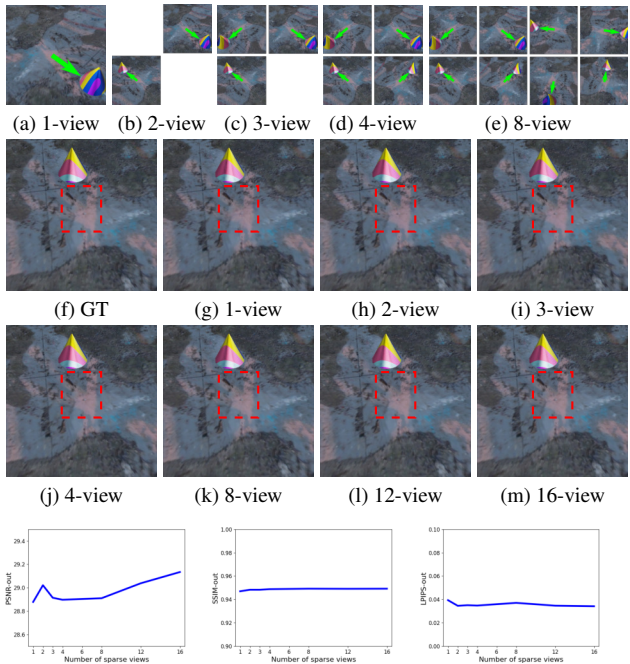

(a) 1-view  (b) 2-view  (c) 3-view  (d) 4-view  (e) 8-view

(f) GT  (g) 1-view  (h) 2-view  (i) 3-view

(j) 4-view  (k) 8-view  (l) 12-view  (m) 16-view

(n) PSNR-out (↑), SSIM-out (↑), LPIPS-out (↓) with different number of input sparse views

Figure 18. Qualitative and quantitative results of our method with different *number* of sparse-view observations on the Cone dataset. Changing the amount of sparse-view supervision doesn't significantly impact the reconstruction quality of object un-occluded regions. (a-e): Sparse-view observations (12- and 16-view cases omitted); (f-m): Evaluation-view renders by the updated NeRF; (n): The evaluation metrics under different number of sparse views.

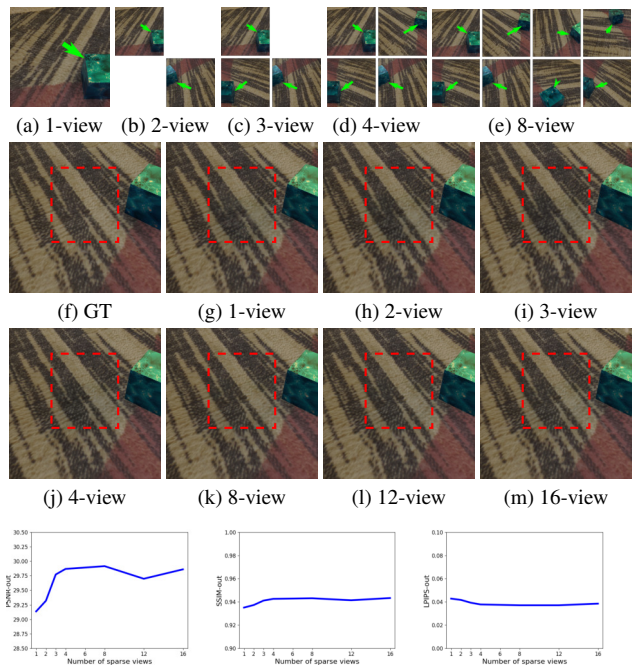

(a) 1-view  (b) 2-view  (c) 3-view  (d) 4-view  (e) 8-view

(f) GT  (g) 1-view  (h) 2-view  (i) 3-view

(j) 4-view  (k) 8-view  (l) 12-view  (m) 16-view

(n) PSNR-out (↑), SSIM-out (↑), LPIPS-out (↓) with different number of input sparse views

Figure 17. Qualitative and quantitative results of our method with different number of sparse-view observations on the Cone dataset. Changing the amount of sparse-view supervision doesn't significantly impact the reconstruction quality of object un-occluded regions. (a-e): Sparse-view observations (12- and 16-view cases omitted); (f-m): Evaluation-view renders by the updated NeRF; (n): The evaluation metrics under different number of sparse views.

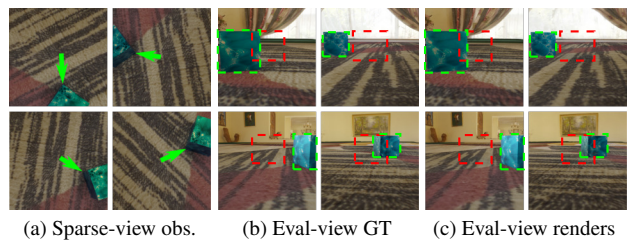

(a) Sparse-view obs.  (b) Eval-view GT  (c) Eval-view renders

Figure 19. Qualitative results of our method on the Cube dataset with different *placements* of sparse view cameras. We capture the changes from the top views of the scene and evaluate at bottom views.

## 12.3. Different placements of sparse-view cameras

Our method doesn't make a strong assumption about the placement of the sparse-view cameras. We have demonstrated that our method can handle sparse views containing overlapped move-in and move-out regions (Fig. 13), and doesn't require a 360deg sparse view coverage around the scene (Fig. 17 (b,c) and Fig. 18 (b,c)). The images can be positioned with less than 180deg coverage around the scene.

Table 8. Quantitative results of our method on the Cube dataset with different *placements* of the sparse view cameras. We capture sparse images from top views of the scene and evaluate at bottom views. We also report the results of evaluating at the same views as in the original experiment. In the move-in region, despite limited observation of the cube's sides, there is no significant drop in the evaluation metrics. Move-out region's reconstruction quality only *slightly* decreases even when evaluated from afar.

| | Methods | PSNR-**in** | SSIM-**in** | LPIPS-**in** | PSNR-**out** | SSIM-**out** | LPIPS-**out** |
|---|---|---|---|---|---|---|---|
| Ours | same view+same eval | **29.96** | **0.95** | **0.04** | 28.45 | **0.97** | **0.02** |
| | top view+same eval | 27.92 | 0.91 | 0.06 | **29.32** | 0.95 | 0.03 |
| | top view+bottom eval | 27.01 | 0.85 | 0.10 | 28.18 | 0.89 | 0.09 |
| | Retrain | 18.46 | 0.68 | 0.39 | 28.76 | 0.96 | 0.03 |

In this section, we show that our method is robust to viewing angles of the sparse cameras. We capture the changes from top views of the scene and evaluate the novel view renders at bottom views, as demonstrated in Fig. 19. The qualitative and quantitative results are reported in Fig. 19 and Tab. 8. In the move-in region, despite limited observation of the cube's sides, there's only a slight impact on evaluation metrics, and the degradation is barely noticeable visually. This suggests our scene change estimation pipeline can effectively cope with reduced sparse-view guidance. For the move-out region, the reconstruction quality only mildly drops even when evaluated from afar.

## 13. Limitations and future work

### 13.1. Occlusion

Under self-occlusion or severe inter-object occlusion, certain object parts may not be visible from the pre-training views. If the object configuration changes, such as revealing the bottom side of an object or removing an object that was causing the occlusion, our method does not reconstruct the un-occluded object parts. A good example is the "Russian doll" scenario: When the biggest doll is moved away, the second biggest doll, which was previously invisible, is exposed. With little prior for this newly revealed doll, our problem reduces to few-view NeRF reconstruction. Thus, this limitation also opens an avenue for future work: integrating a few-view NeRF method (e.g., [55]) into our pipeline could potentially address this challenge.

### 13.2. Complex lighting effects

The environmental lighting in the scene may not be consistent across all regions. If an object's pose changes, its lighting can change significantly. Our method does not yet account for the variations in lighting effects. After transforming the 3D coordinates for the pre-trained NeRF, the object shadow, reflection, etc. might be very different from their original state. As a future direction, a NeRF relighting method [19, 39, 52] can be integrated into our pipeline to enhance the object appearance consistency.

### 13.3. Non-rigid object reconfigurations

Our method uses 6DoF transformation matrices to represent the object pose changes, which restricts its immediate application to non-rigid object reconfiguration scenarios. However, this is *not a fundamental limitation* of our method, since the other components of our pipeline, apart from the pose change estimation module, does not rely on the assumption of object rigidity. We can readily replace the rigid object pose change estimation method with its non-rigid counterparts (e.g., [4]) to accommodate these changes.